%% file: Main.tex
\documentclass[10pt,twocolumn,letterpaper]{article}

\usepackage{cvpr}              

\usepackage{graphicx}
\usepackage{float}
\usepackage{multirow} 
\usepackage{booktabs}
\usepackage{times}
\usepackage{epsfig}
\usepackage{amsmath}
\usepackage{amssymb}
\usepackage{url}
\usepackage{makecell}

\usepackage[accsupp]{axessibility}  


\definecolor{cvprblue}{rgb}{0.21,0.49,0.74}
\usepackage[pagebackref,breaklinks,colorlinks,allcolors=cvprblue]{hyperref}

\def\confName{CVPR}
\def\confYear{2025}

\title{MVD-HuGaS: Human Gaussians  from a  Single Image via \\ 
3D Human Multi-view Diffusion Prior}

\author{
Kaiqiang Xiong\textsuperscript{1,2} \quad 
Ying Feng\textsuperscript{4} \quad 
Qi Zhang\textsuperscript{5} \quad 
Jianbo Jiao\textsuperscript{6} \quad
Yang Zhao\textsuperscript{7}  \\
Zhihao Liang\textsuperscript{8} \quad  
Huachen Gao\textsuperscript{1} \quad 
Ronggang Wang\textsuperscript{1,2,3}\\
\textsuperscript{1}School of Electronic and Computer Engineering, Peking University\\
\textsuperscript{2}Peng Cheng Laboratory\quad
\textsuperscript{3}Migu Culture Technology Co., Ltd\\
\textsuperscript{4}vivo Mobile Communication (Hangzhou) Co., Ltd. \quad
\textsuperscript{5}vivo Mobile Communication Co. Ltd\\
\textsuperscript{6}School of Computer Science, University of Birmingham\\
\textsuperscript{7}School of Computer and Information, Hefei University of Technology\\
\textsuperscript{8}South China University of Technology \quad 
 \\
{\tt\small xiongkaiqiang@stu.pku.edu.cn \quad rgwang@pkusz.edu.cn}
}

\begin{document}

\makeatletter
\let\@oldmaketitle\@maketitle
\renewcommand{\@maketitle}{\@oldmaketitle

\resizebox{1.0\linewidth}{!}{
\includegraphics[trim={0cm 0cm 0cm 0cm},clip, width=1\linewidth]{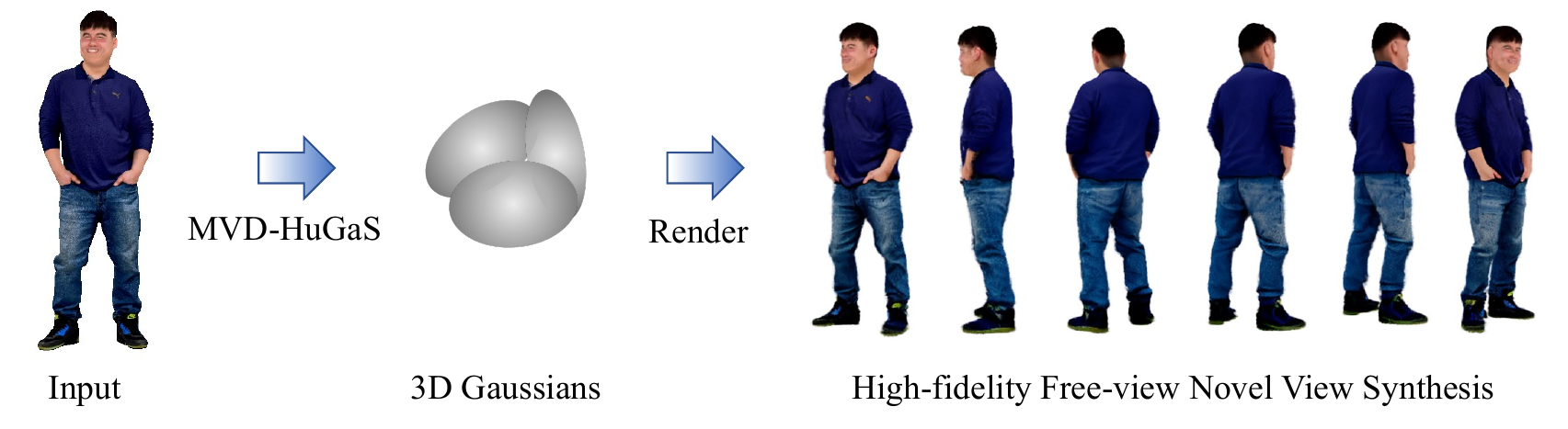}
}

\centering
\captionsetup{hypcap=false}
\captionof{figure}{
Given a single image, the proposed MVD-HuGaS generates 3D Gaussians enabling high-fidelity free-view 3D clothed human Novel View Synthesis.} 
\vspace{10pt}

\label{fig:head}
}

\makeatother

\maketitle

\begin{abstract}
3D human reconstruction from a single image is a challenging problem and has been exclusively studied in the literature.
Recently, some methods have resorted to diffusion models for guidance, optimizing a 3D representation via Score Distillation Sampling(SDS) or generating one back-view image for facilitating reconstruction. 
However, these methods tend to produce unsatisfactory artifacts (\textit{e.g.} flattened human structure or over-smoothing results caused by inconsistent priors from multiple views) and struggle with real-world generalization in the wild.
In this work, we present \emph{MVD-HuGaS}, enabling free-view 3D human rendering from a single image via a multi-view human diffusion model.
We first generate multi-view images from the single reference image with an enhanced multi-view diffusion model, which is well fine-tuned on high-quality 3D human datasets to incorporate 3D geometry priors and human structure priors.
To infer accurate camera poses from the sparse generated multi-view images for reconstruction,  an alignment module is introduced to facilitate joint optimization of 3D Gaussians and camera poses.
Furthermore, we propose a depth-based Facial Distortion Mitigation module to refine the generated facial regions, thereby improving the overall fidelity of the reconstruction.
Finally, leveraging the refined multi-view images, along with their accurate camera poses, MVD-HuGaS optimizes the 3D Gaussians of the target human for high-fidelity free-view renderings.
Extensive experiments on Thuman2.0 and 2K2K datasets show that the proposed MVD-HuGaS achieves state-of-the-art performance on single-view 3D human rendering.
\end{abstract}

\section{Introduction}

Reconstructing the 3D representation of a clothed human from only a single-view image is a fundamental challenge in 3D computer vision and computer graphics, extensively investigated due to its broad applicability in industries such as VR/AR, video gaming, and movie production.
The inference of a 3D structure of the visible regions from a single image is extremely ill-posed, while the estimation of both geometry and appearance of the residual invisible regions further exacerbates this problem.

To address these challenges, a surge of learning-based methods~\cite{saito2019pifu,chan2024fine,saito2020pifuhd,zheng2021pamir,huang2024tech,alldieck2019tex2shape,alldieck2022photorealistic,hu2023sherf,huang2023one,albahar2023single,zhu2019detailed,jiang2020bcnet,he2020geo,xiu2022icon,huang2020arch,he2021arch++,liao2023high} has been developed to advance the effectiveness of single-view 3D human reconstruction.
Previous methods \cite{saito2019pifu,saito2020pifuhd,zheng2021pamir} train 3D generative models to predict 3D neural fields from pixel-aligned features.
However, these methods often fail to generalize to various unseen scenarios due to the limited size of 3D human datasets \cite{yu2021function4d,han2023high,liao2023high,cai2022humman}.

Recently, motivated by the success of generic image-to-3D, several methods~\cite{huang2024tech,ho2024sith,albahar2023single} turn to 2D image diffusion models for guidance. 
Owing to pretraining on the Internet-scale image datasets, 2D diffusion models, such as Stable Diffusion~\cite{rombach2022high} and Imagen~\cite{saharia2022photorealistic}, have acquired a prior that encapsulates the natural distribution of images.
Moreover, camera-conditioned diffusion models, including single-view \cite{liu2023zero} and multi-view variants \cite{liu2023syncdreamer, voleti2024sv3d}, incorporate camera embedding conditions to generate images on specified viewpoints, which further promotes the lifting from 2D images to 3D.
With a fine-tuned single-view diffusion model, \cite{ho2024sith} generates one back-view image via an image-to-image diffusion model, followed by a 3D human reconstruction process via the Skinned Multi-Person Linear model (SMPL) \cite{loper2023smpl,feng2021collaborative,pavlakos2019expressive}.
Nevertheless, the information provided by just two images is insufficient for accurate 3D human reconstruction, resulting in a flattened appearance on the side view.
Besides, \cite{huang2024tech,zhang2024humanref} apply the SDS to optimize a 3D human presentation.
However, given an image as a condition, single-view diffusion models cannot guarantee 3D geometric consistency among the generated images, 
which leads to over-smoothing textures.
In essence, existing methods \cite{zhang2024sifu, ho2024sith} face significant challenges in achieving high-fidelity novel view synthesis for a 3D human from a single image.
To this end, we propose to leverage the multi-view diffusion prior for single-view 3D human reconstruction.

\begin{figure}[t]
  \vspace{-0.2cm}
  \begin{center}
     \includegraphics[trim={0cm 0.3cm 0cm 0cm},clip,width=1.0 \linewidth]{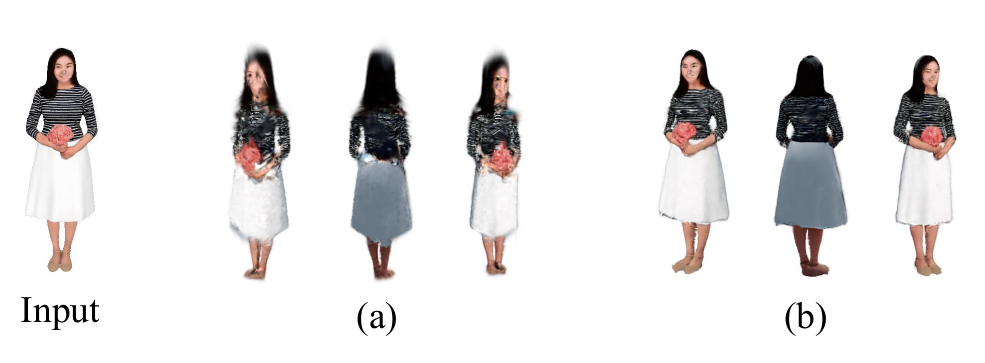}
  \end{center}
  \vspace{-0.4cm}
  \caption{\textbf{Camera Misalignment Issue.} (a) 3D-GS rendering results from generated multi-view images with conditioned camera poses. (b) 3D-GS rendering results from multi-view images with the optimized camera poses via our Camera Alignment Module.}
  \label{fig:nopose}
  \vspace{-0.4cm}
\end{figure}

In contrast to single-view diffusion models~\cite{liu2023zero, liu2024one} that generate multi-view images through multiple iterations, suffering from multi-view inconsistency,
recent multi-view diffusion models, such as Stable Video 3D (SV3D)~\cite{voleti2024sv3d} and SyncDreamer~\cite{liu2023syncdreamer}, leverage cross-attention mechanisms to facilitate the multi-view information interaction during denoising and become capable of producing more geometrically consistent multi-view images through a single inference.
They have also been demonstrated to achieve remarkable results in generic 3D object generation tasks.

\begin{figure}[t]
  \vspace{-0.2cm}
  \begin{center}
     \includegraphics[trim={0cm 0.3cm 0cm 0cm},clip,width=1.0 \linewidth]{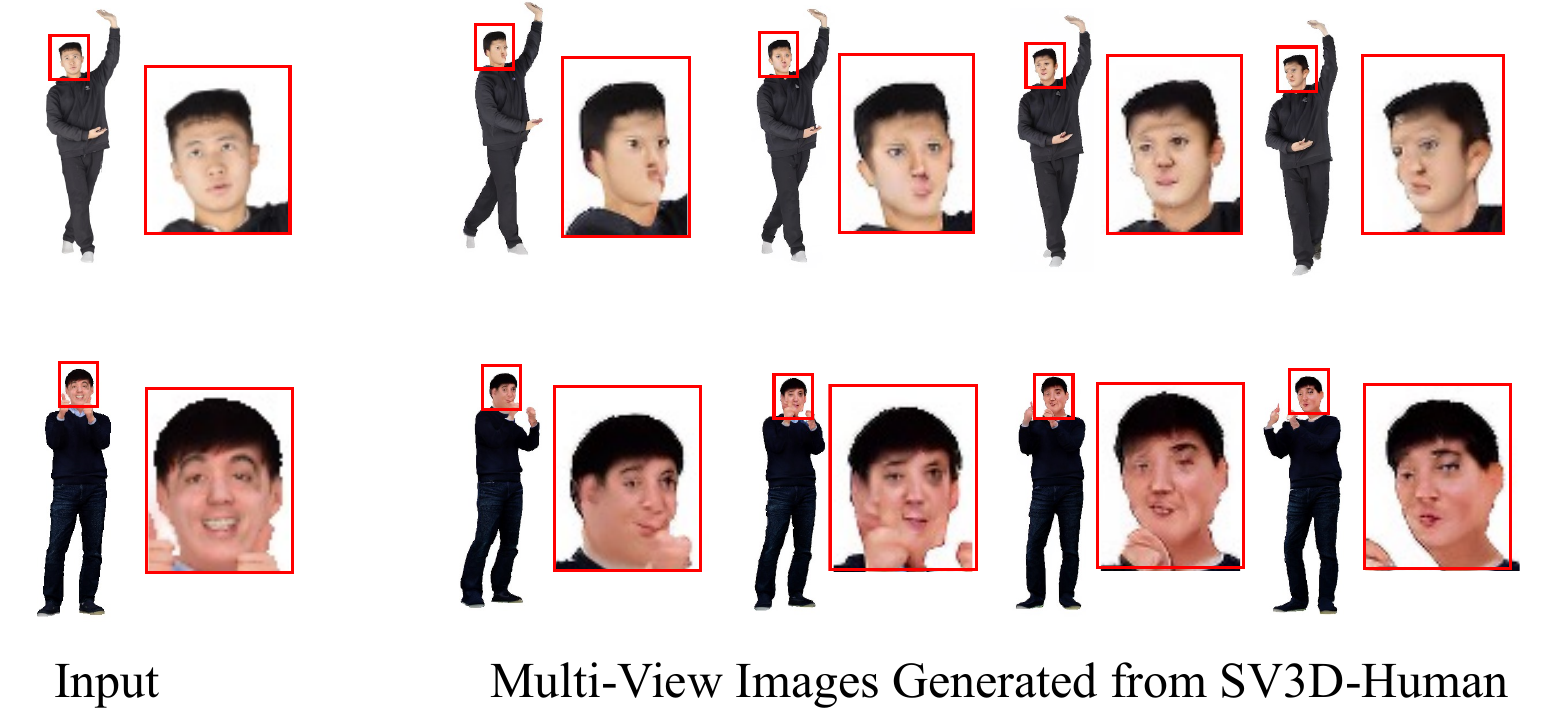}
  \end{center}
  \vspace{-0.4cm}
  \caption{\textbf{Facial Distortion Issue.} Despite being fine-tuned on 3D human datasets, models still struggle to generate multi-view consistent and plausible human faces, even though they can learn the structural priors of the human body.}
  \label{fig:face_distoration}
  \vspace{-0.4cm}
\end{figure}

However, extending generic multi-view diffusion models \cite{long2024wonder3d, voleti2024sv3d, liu2024one1} to 3D human reconstruction presents several challenges:
\textbf{(1) 3D Human Body Priors Deficiency.} These models are often trained on datasets that lack human data, resulting in a deficiency of essential knowledge about human anatomy (\textit{i.e.}, structure) and appearance. Roughly applying the priors from these models indeed results in unrealistic human depictions.
\textbf{(2) Camera Misalignment.} To our knowledge, all camera-conditioned diffusion models \cite{liu2024one, liu2024one1, shi2023zero123++}, including SV3D, cannot ensure perfect alignment with specified camera parameters, leading to mismatches between the camera parameters and the generated images, thus affecting the quality of 3D human reconstruction, as depicted in \cref{fig:nopose}.
\textbf{(3) Facial Distortion.} 
These models still struggle to generate accurate human faces, even when fine-tuned on 3D human datasets, as \cref{fig:face_distoration} shows. 
The dimensionality reduction in latent diffusion models may lead to significant facial detail loss, especially since faces occupy a small image region. 
Moreover, the human visual system is sensitive to facial details, which makes the removal of facial distortion even more necessary and challenging.

To address these challenges, this paper proposes \textbf{MVD-HuGaS}, a pipeline that consists of the following modules:
\textbf{SV3D-Human:}
We enhance SV3D with a high-quality 3D human dataset \cite{han2023high}, instilling in the model a comprehensive understanding of human structure knowledge. This adaptation enables the generation of realistic multi-view human images.
\textbf{Camera Alignment Module:}
We use 3D Gaussians~\cite{kerbl20233d} to represent the 3D human and simultaneously optimize the 3D human with learnable camera poses using the image reconstruction loss.
This scheme effectively corrects camera poses in the generated multi-view images, which are almost sparse and exhibit pixel-level geometric inconsistencies.
\textbf{Facial Distortion Mitigation:}
We devise a depth-based warping module to enhance the fidelity of facial regions. 
Specifically, utilizing a multi-view 3D Morphable Model (3DMM) fitting \cite{blanz2023morphable}, we extract a coherent reference depth map from the inconsistent and distorted generated facial parts. 
Subsequently, a depth-guided forward-warping technique is used to warp the reference face onto the source views and effectively restore them.
This scheme incorporates facial geometric priors with the input texture (\textit{i.e.} reference image) to mitigate the mentioned facial distortions.
\textbf{Final Reconstruction with 3D Gaussians:}
After obtaining the aligned cameras and improved faces, we reuse the 3D Gaussians to represent and reconstruct the target 3D human.
Benefiting from this representation, MVD-HuGaS achieves high-fidelity and real-time rendering performance from arbitrary viewpoints, as depicted in \cref{fig:head}.

In summary, the main contributions of this paper include:
\begin{itemize}

\item A single-view 3D human reconstruction pipeline, \textit{MVD-HuGaS}, is proposed, enabling free-view high-fidelity rendering. The state-of-the-art (SOTA) performance on benchmarks validates its effectiveness.

\item SV3D-Human produces realistic multi-view human images after fine-tuning on high-quality 3D human datasets, which supports subsequent 3D human reconstruction.

\item We embed the camera misalignment problem in previous camera-conditioned diffusion models and propose a camera alignment module to obtain accurate camera poses from the sparse generated multi-view images.

\item We devise a depth-based facial distortion mitigation module, combining human face geometry priors and input texture information to inpaint the generated distorted faces.

\end{itemize}

\begin{figure*}[htbp]
  \vspace{-0.5cm}
  \begin{center}
     \includegraphics[trim={0cm 0cm 0cm 0cm},clip,width=1.0\linewidth]{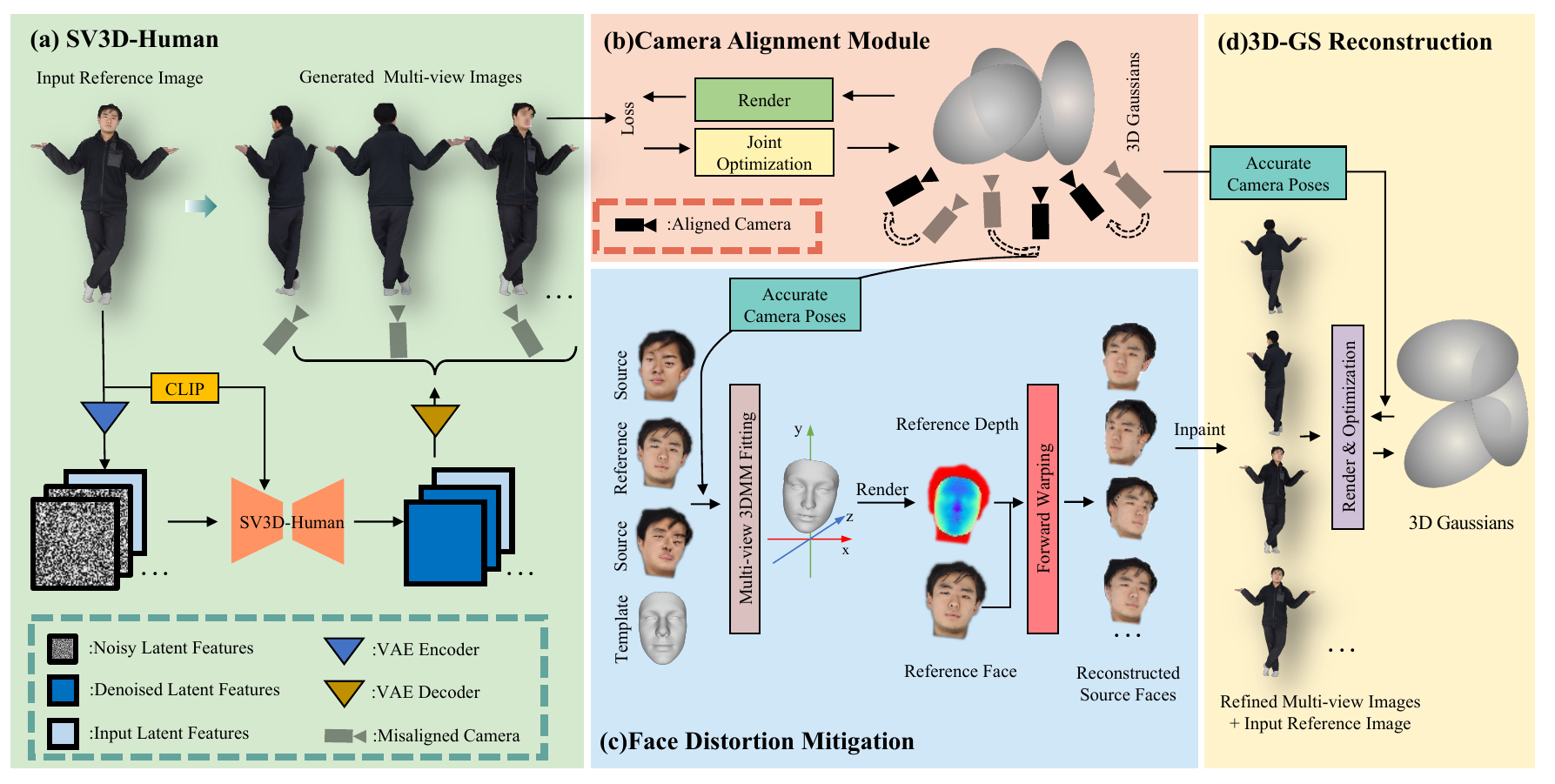}
  \end{center}
  \vspace{-0.3cm}
  \caption{\textbf{Framework of MVD-HuGaS.} The framework consists of four modules:
\textbf{(a) SV3D-Human:} Starting with an input reference image, our SV3D-Human model initiates the process by generating multi-view images. These images are produced with misaligned camera poses and facial distortions.
\textbf{(b) Camera Alignment Module: }Our camera alignment module works to jointly optimize the 3D Gaussians and the misaligned camera poses. This is achieved by minimizing the reconstruction loss between the rendered 3D Gaussians and the multi-view images. After optimization, we obtain camera poses that are precisely aligned with the generated multi-view images.
\textbf{(c) Face Distortion Mitigation: }To mitigate the facial distortion in the generated source images, we first apply the aligned camera poses and conduct multi-view 3D Morphable Model (3DMM) fitting. This process yields a human face mesh with accurate shape and location in 3D space. The mesh is then rendered back to the reference view to derive the depth of the reference face. Utilizing the reference RGB image and its corresponding depth map, the faces in the source views are then reconstructed through depth-based forward warping.
\textbf{(d) 3D-GS Reconstruction: }Armed with the accurate camera poses from step (b), the refined multi-view images from step (c), and the input reference image, we reuse the 3D Gaussians to present and reconstruct the target 3D human, enabling high-fidelity real-time free-view renderings.}
  \label{fig:pipeline}
  \vspace{-0.4cm}
\end{figure*}

\section{Related Work}
\noindent\textbf{Single-view 3D Human Reconstruction}
Since the inherent ambiguity in the monocular geometry estimation, reconstructing a 3D human from a single image is challenging.
Besides, the geometry and appearance of invisible areas are entirely unknown, introducing more complexity.
Recently, PIFu~\cite{saito2019pifu} proposes to obtain pixel-level features from 2D images, then transform the features into a 3D geometric and texture field, and finally use the Marching Cubes algorithm to extract the mesh.
PIFuHD~\cite{saito2020pifuhd} incorporates high-resolution features to introduce more details.
While a number of methods~\cite{zhang2024sifu, zheng2021pamir, ho2024sith, zhang2024humanref, xiu2023econ, xiu2022icon, huang2023one, albahar2023single} combine parametric models with estimated normal or introduce additional priors from diffusion models~\cite{rombach2022high} to enhance geometric quality, they fail to produce satisfactory appearance.
To overcome this limitation, PHORHUM~\cite{alldieck2022photorealistic} and S3F~\cite{corona2023structured} decompose the appearance into albedo and illumination.
For diversity, Context-Human~\cite{gao2024contex} and Tech~\cite{huang2024tech} integrate the SDS~\cite{poole2022dreamfusion} approach, but they still suffer from multi-view inconsistency and saturation.
In this work, we leverage the 3D awareness of the SV3D~\cite{voleti2024sv3d} model and infuse it with human awareness to obtain SV3D-Human.
With the guidance of the human-aware multi-view diffusion models, more coherent multi-view images can be produced for subsequent 3D human reconstruction.

\noindent\textbf{Image-to-3D}
Recent 3D generation methods resort to diffusion models to reconstruct general 3D objects from a single image.
As a pioneer, Zero123~\cite{liu2023zero} proposes an algorithm for generating novel-view images from a single input image with the given camera conditions.
Based on Zero123, One-2-3-45~\cite{liu2024one} inputs the generated multi-view images into a feed-forward network and achieves rapid 3D reconstruction.
However, the multi-view images generated by Zero123 lack multi-view consistency and the reconstruction results suffer from poor quality.
Subsequent methods like SyncDreamer~\cite{liu2023syncdreamer}, Zero123++~\cite{shi2023zero123++}, One-2-3-45++~\cite{liu2024one1}, Consistent123~\cite{weng2023consistent123}, and Wonder3D ~\cite{long2024wonder3d} improve the consistency of generated multi-view images by employing attention mechanisms for information exchange during denoising. 
Recently, SV3D~\cite{voleti2024sv3d} fine-tuned the Stable Video Diffusion (SVD)~\cite{blattmann2023stable} model on large-scale 3D datasets~\cite{deitke2023objaverse}.
Benefiting from the internal consistency of multiple frames rendered using 3D assets, SV3D enhances the multi-view consistency.
However, directly applying current multi-view generation methods to 3D human generation faces two challenges.
Firstly, despite the camera parameters being used as conditions, the generated images indeed mismatch the given parameters.
Although several methods have improved the multi-view consistency, there is still a huge gap compared to real camera-captured images.
Due to the sparsity of the generated images and the lack of perfect pixel-level consistency, previous calibration methods (\textit{e.g.} COLMAP~\cite{schonberger2016structure, barnes2009patchmatch}, Flowmap~\cite{smith2024flowmap}, and DUSt3R~\cite{wang2024dust3r}) fail to recover camera poses under these conditions.
Secondly, owing to the intricate structure of the human face and the inevitable information loss from the VAE~\cite{kingma2013auto}, there remains a challenge in generating fine facial regions, which is a critical aspect of high-quality 3D human reconstruction.
In this work, the proposed MVD-HuGaS handles these issues to facilitate 3D human reconstruction from one single image.

\section{Method}
Given a frontal image of a clothed human, the proposed MVD-HuGaS reconstructs a 3D Human representation that supports free-viewpoint rendering.
The proposed pipeline mainly consists of four stages, as summarized in \cref{fig:pipeline}:
Firstly, our SV3D-Human generates multi-view images from a single reference image.
Secondly, we introduce a camera alignment module to recover accurate camera poses from the sparse generated images.
Subsequently, a facial distortion mitigation module is devised to inpaint the distorted facial regions in the generated images.
Finally, with the camera-aligned and face-inpainted multi-view images, MVD-HuGaS reconstructs the final 3D human Gaussians, enabling real-time novel view synthesis.

\subsection{3D-Human-Aware Multi-view Diffusion Model}
\noindent \textbf{Multi-view Diffusion Model SV3D}
Given the reference image $I_0$, the pretrained SV3D-u ~\cite{voleti2024sv3d} generates a sequence of video frames $\{I_i|i \in \{ 1, ..., N  \} \}$ of the target object in $I_0$.
Assuming that the object is at the center of the canonical space, the corresponding camera poses can be denoted as $\{$azimuth $a$, elevation $e$, radius $r$ $\}$.
The ideal generated $N$ frames match the input's elevation and radius $\{e_i=e_0, r_i=r_0 |i\in \{ 1, ..., N  \} \}$, and their azimuths $\{a_i = i / (N + 1) * 2 \pi|i \in \{ 1, ..., N  \} \}$ are evenly distributed.

While the pretrained SV3D model effectively produces multi-view images of static 3D objects,
directly applying it to 3D human reconstruction results in a significant performance decline.
This is partly due to that the model lacks 3D human understanding, as it is primarily trained on datasets devoid of 3D humans. 
The complexity of the human body, with its non-rigid movements, occlusions, and detailed features like faces and hands, exceeds the pretraining scope of the model. 
To address this, we introduce \emph{SV3D-Human}, a model infused with 3D human knowledge for enhanced human-aware reconstruction capabilities.

\noindent\textbf{SV3D-Human}
A straightforward way is to integrate SMPL-predicted meshes into the diffusion model for additional priors. 
However, inaccuracies in SMPL parameter estimation can introduce fatal errors that distort the diffusion and subsequent reconstruction, limiting generalization across diverse human poses.
Drawing from success in camera-conditioned models that fine-tune on 3D datasets to gain 3D awareness, we fine-tune SV3D on a high-quality 3D human dataset, 2K2K~\cite{han2023high}, to imbue it with a deeper understanding of 3D human structure.
The 2K2K dataset comprises over two thousand high-fidelity 3D clothed human models, each with one million vertices and corresponding texture maps. 
The dataset diversity, with scans featuring different individuals, is ideal for injecting human awareness into the multi-view diffusion model.

Specifically, for fine-tuning, each 2K2K model is rendered into $N+1$ uniformly spaced images at a fixed elevation and radius.
During fine-tuning, a VAE encodes $N+1$ images into latent features, and noise is added to $N$ target features using the EDM schedule~\cite{karras2022elucidating}.
Given the input latent code, noisy target latent codes, camera pose embedding, and CLIP embedding~\cite{radford2021learning}, the model learns to predict noise strength under $L2$ loss supervision.
Post-finetuning, SV3D-Human emerges, integrating vast 2D images, 3D geometric, and 3D human structure priors. 
With a single human image, SV3D-Human can generate coherent multi-view images around the central 3D human.

\begin{figure*}[h]
  \vspace{-0.3cm}
  \begin{center}
     \includegraphics[trim={0cm 0cm 0cm 0cm},clip,width=0.9 \linewidth]{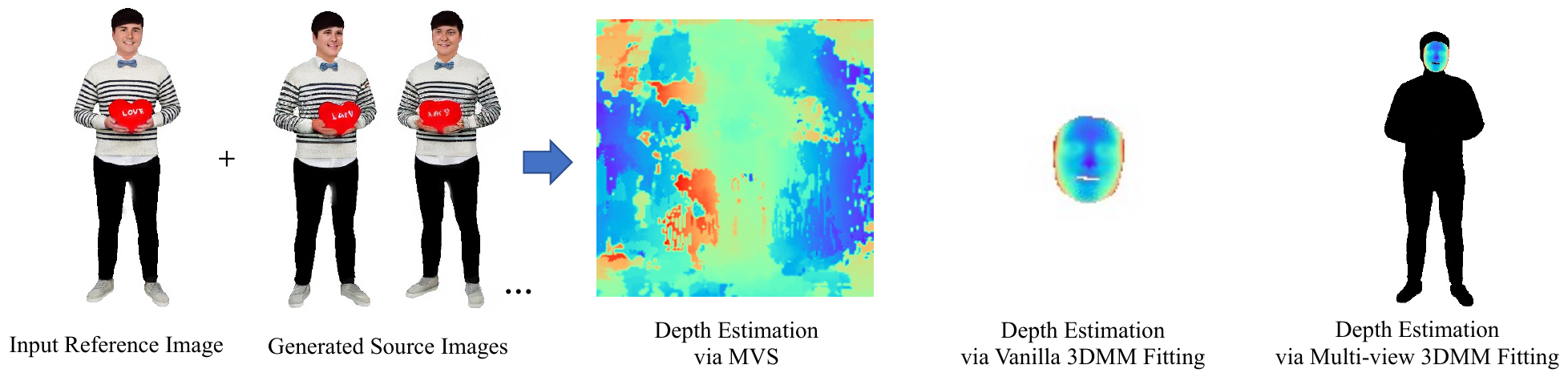}
  \end{center}
  \vspace{-0.4cm}
  \caption{\textbf{Different Depth Estimation Comparison.} MVS methods infer noisy depth maps due to the inherent inconsistency and distortion of generated multi-view face parts. 
  Vanilla 3DMM fitting methods can only estimate face geometry and relative depth information from a single image, but they often fail to accurately capture the absolute spatial positions.
  In contrast, the proposed Multi-view 3DMM Fitting can accurately estimate the absolute depth of the face, including both facial geometry and 3D spatial information.  }
  \label{fig:depth}
  \vspace{-0.4cm}
\end{figure*}
\subsection{Camera Alignment Module}
\noindent\textbf{3D Gaussians Splatting}
3D Gaussians Splatting~\cite{kerbl20233d} is a recently proposed explicit scene representation that supports high-fidelity real-time renderings.
It represents the 3D scene as a set of 3D Gaussians $Gs=\{ G_j| j\in \{1, ..., M \} \}$ and renders images via splitting: 
\vspace{-0.1cm}
\begin{align}
 I^r_{i} =  \Psi(Gs, P_i, K_i) ,
 \label{eq:rendering}
\end{align}
\vspace{-0.1cm}
where $I^r_{i}$ is the rendered image and $P_i, K_i$ are the corresponding camera extrinsics and intrinsic.
Each 3D Gaussian $G_j$ is characterized as $\{ \mu, \Omega, \alpha, F \}$.
$\mu$ is the mean of 3D Gaussian distribution. 
$\Omega$ is the 3D covariance matrix, which can be decoupled as a rotation matrix $R$  and a scaling matrix $S$ for easier optimization. 
Besides, $\alpha, F$  are opacity and spherical harmonics coefficients for rendering.
During rendering, Guassians are projected to the 2D plane via viewing transformation $W$ and the Jacobian of the affine approximation $J$ of the projection transformation:
$\Omega ' = JW \Omega W^{T} J^{T}.$
The final color $C$ of each pixel is acquired in an alpha-blending way according to the depth order of the $O$ overlapping Gaussians:
$C = \Sigma_{i \in O} \alpha_{i} c_{i} \prod_{j=1}^{i-1}(1-\alpha_j)$.
A tile-based rasterizer is used for efficient rendering.
During optimization, the Gaussian parameters are updated under the reconstruction loss and SSIM loss~\cite{wang2004image} between rendered and ground-truth images:
\vspace{-0.1cm}
\begin{align}
L = (1-\lambda)L_1 + \lambda L_\text{SSIM},
\label{eq:gs_loss}
\end{align}
\vspace{-0.1cm}
where $\lambda$ is the weighting parameter.
More details can be found in the~\cite{kerbl20233d}.
Although 3D Gaussians achieve high-quality real-time rendering, one limitation of 3D Gaussians is their requirement for \emph{accurate camera poses} to learn scene representations.

\noindent\textbf{Joint Optimization of Camera Poses and 3D Gaussians} 
As aforementioned, camera-conditioned diffusion models, including SV3D, commonly assume ideal alignment between generated images with camera conditions, an expectation that is frequently unmet due to imperfect image generation. 
Our SV3D-Human, inheriting from SV3D, encounters a similar challenge. 
During generation, it implicitly infers elevation from the input single image for multi-view generation but with inaccuracies. 
Due to the intricate nature of the 3D human, camera misalignment can significantly impair the quality of subsequent 3D reconstructions.

To handle this issue, we introduce a joint optimization strategy.
\cite{lin2021barf, bian2023nope} have demonstrated that the camera parameters and a neural radiance field (NeRF) can be predicted simultaneously by minimizing the reconstruction loss between the rendered images and the ground truth images.
Considering that NeRF is particularly sensitive to sparse, low-textured, and geometrically inconsistent views, we devise an optimization strategy to jointly optimize the camera poses and the 3D Gaussians.
Mathematically, this joint optimization can be formulated as:
\vspace{-0.1cm}
\begin{align}
Gs ^\ast, P ^\ast = \sum \limits_{i=0} \limits^{N} \arg \min \limits_{Gs, P} L(I_i, I_i^r ),
\end{align}
\vspace{-0.1cm}
where $L$ is referred in \cref{eq:gs_loss},  $\{I_i|i \in \{ 0, 1,..., N  \} \}$ is the multi-view images, including the generated images and the input image.
The essence of our approach is rendering the scene as a function of 3D Gaussians and variable camera parameters in \cref{eq:rendering}, then simultaneously updating both the 3D Gaussian and camera parameters based on the rendering loss in \cref{eq:gs_loss}.
Note that our focus is solely on optimizing the camera's extrinsic parameters. 
We initiate the process with the ideal camera poses, leveraging the explicit representation of 3D Gaussians to quickly and accurately determine the unknown extrinsic parameters.

\subsection{Face Distortion Mitigation}
Despite obtaining accurate camera parameters, the recovery of high-fidelity 3D humans still faces facial distortion issues.
The facial region, more intricate and subtle than the body, contains detailed features within a small region, making it highly sensitive to visual perception. 
The facial distortions can markedly affect human visual perception. 
Concurrently, the generation of realistic faces is a considerable hurdle for latent diffusion models, with the potential for information loss during the VAE encoding into latent space, often leading to unavoidable facial distortions.
To address this, MVD-HuGaS integrates the appearance information in the input image to mitigate the distortions of generated faces.

\begin{figure*}[t]
  \vspace{-0.4cm}
  \begin{center}
     \includegraphics[trim={0cm 0cm 0cm 0cm},clip,width=0.9\linewidth]{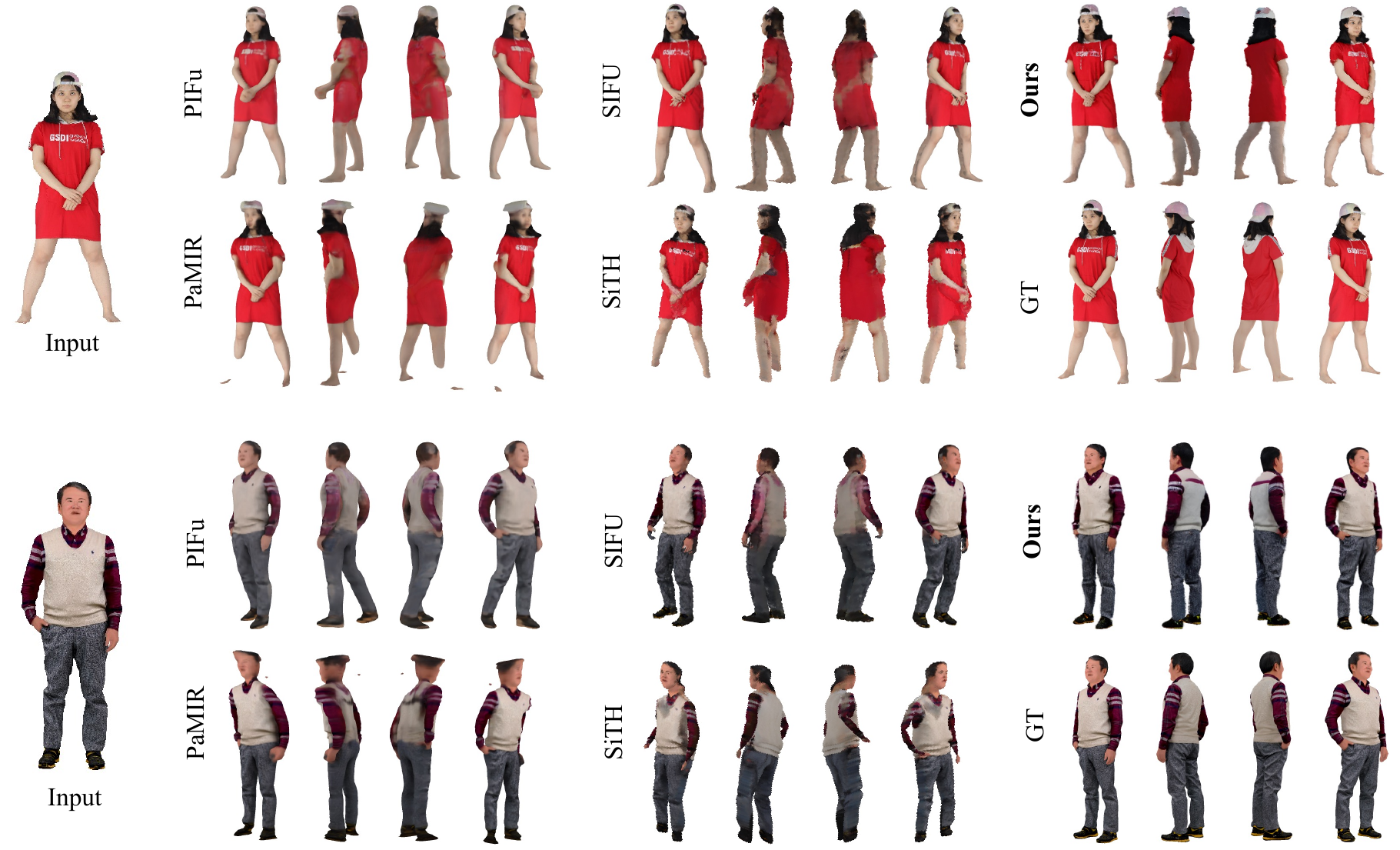}
  \end{center}
  \vspace{-0.3cm}
  \caption{\textbf{Qualitative comparison on Thuman2.0 and 2K2K.} More results can be found in the supplementary file. }
  \label{fig:qualitative_comparison}
  \vspace{-0.2cm}
\end{figure*}

\noindent\textbf{Depth Estimation with Mutli-view 3DMM Fitting}
\emph{Multi-view Stereo (MVS)} \cite{yao2018mvsnet} estimates depth maps from calibrated RGB images using plane sweeping and feature matching.
With a reference image $I_0$ and corresponding source images $\{I_i| i=1,2,3,...\}$, it predicts the depth map for the reference image while relying on the ideal assumption of photometric consistency -- 3D points maintaining a consistent appearance across views.
However, this assumption can not be satisfied in generated images, especially in facial regions prone to distortion.
\emph{3D morphable models (3DMMs)}~\cite{blanz2023morphable, cao2013facewarehouse}, parametric representations of human faces, facilitate the alignment of a template mesh to a facial image through optimization. 
This process leverages Principal Component Analysis (PCA) to simplify prediction by reducing dimensionality. 
However, 3DMMs can not capture the photorealistic appearance of real human faces, and the facial mesh derived from a single image almost lacks precise 3D spatial information, as shown in \cref{fig:depth}.

To this end, we introduce a multi-view 3DMM fitting approach that leverages the advantages of both MVS and 3DMM.
We aim to predict accurate facial depth $D^f_0$ from the reference image and two generated source images $\{I_i|i=0,1,2\}$.
Initially, we segment the facial regions $\{I^f_i|i=0,1,2\}$ and estimate corresponding landmarks $\{LM_i|i=0,1,2\}$ across the images. 
During optimization, the facial mesh is rendered onto the 2D plane of the images with aligned camera poses to acquire rendered 2D face images $\{I^{fr}_i|i=0,1,2\}$ and rendered 2D landmarks $\{LM^r_i|i=0,1,2\}$ via PyTorch3D.
The 3DMM parameters, including shape, 3D location, and rotation, are refined by minimizing a composite loss function:
\vspace{-0.1cm}
\begin{align}
L_{3DMM}  = \lambda_{img} L_{img} + \lambda_{lm} L_{lm} + \lambda_{reg} L_{reg},
\end{align}
\vspace{-0.1cm}
where $L_{img}$, $L_{lm}$, and $L_{reg}$ denote the image reconstruction loss, landmark loss, and regularization loss, respectively~\cite{guo20213d}. 
After optimization, we obtain an accurate facial mesh from the initially distorted faces.
We then render this final facial mesh into the reference view to derive the reference facial depth $D^f_0$.

\noindent\textbf{Forward Warping for Face Restoration}
We advocate for an approach that merges geometric insights from a fitted 3DMM mesh with the visual details from the input reference image.
Given a reference facial image $I^f_0$, its depth map $D^f_0$, associated intrinsic $K_0$, and relative transformation $T$, the reconstructed source facial images $\{\hat I^f_i|i=1,2,... \}$  can be obtained through forward warping.
For a specific pixel $p$ in $I^f_0$, its coordinate $p'$ in a source image $\hat I^f_i$ can be calculated:
\vspace{-0.1cm}
\begin{align}
p'=KT(D^f_0(p) \cdot K_0^{-1}p).
\end{align}
\vspace{-0.1cm}
Then the generated images are enhanced by inpainting with the reconstructed source facial images $\{\hat I^f_i|i=1,2,... \}$, resulting in the refined multi-view images $\{\hat I_i|i=1,2,... \}$.


\subsection{High-fidelity Real-time Free-view Rendering}

\noindent \textbf{Fast Initialization with MVS}
With the accurate camera poses and inpainted multi-view images, we employ a learned MVS method~\cite{peng2022rethinking} for fast initialization.
The process involves feeding multi-view images and their corresponding source images into the MVS network to yield multi-view depth maps.
Afterwards, a quick filtering phase ensures photo consistency and geometric consistency, eliminating noisy pixels.
The filtered points are then integrated into a global 3D space, culminating in a sparse 3D point cloud that serves as the 3D Gaussian initialization.

\noindent \textbf{Floaters Removing \& Anti-Alias}
There exist floaters near the human body in the initial 3D Gaussian reconstructions, stemming from the white background parts in the input multi-view images. 
To address this, a random background strategy is implemented to eliminate these undesired white spots, ensuring a clean reconstruction of the human body.
Additionally, strong artifacts emerge when adjusting the focal length or camera distance, a common issue attributed to the absence of 3D frequency constraints and the use of 2D dilation filters. 
To counter this, a 3D smoothing filter and a 2D Mip filter are integrated as proposed in~\cite{yu2024mip}, which significantly reduce aliasing artifacts in renderings.

After a brief optimization, the resulting 3D Gaussians enable high-fidelity real-time rendering of 3D humans.

\section{Experiments}

\begin{table}[t]
  \vspace{-0.3cm}
  
      \renewcommand\arraystretch{0.8}

  \centering
  \resizebox{1.0\linewidth}{!}{
  \begin{tabular}{l| ccc c}
  \toprule


Method   &  PSNR $\uparrow$ & SSIM $\uparrow$	& LPIPS $\downarrow$ & CLIP-Similarity $\uparrow$  \\

  \midrule
    PIFu \cite{saito2019pifu}              & 18.42   & 0.91	  & 0.09   & 0.80   \\
    PAMIR \cite{zheng2021pamir}            & 18.62   & 0.91	  & 0.09   & 0.81   \\
    SiTH \cite{ho2024sith}              & 16.75 	& 0.89	  & 0.11   & 0.84   \\
    SIFU \cite{zhang2024sifu}             & 16.44 	& 0.89	  & 0.12   & 0.84   \\
    One2345 \cite{liu2024one1}             & 15.20 	& 0.83	  & 0.19   & 0.77   \\
    SV3D \cite{voleti2024sv3d}             & 17.32 	& 0.86	  & 0.14   & 0.83   \\

  \midrule
  \textbf{MVD-HuGaS (Ours)}  	& \textbf{19.02} & \textbf{0.93} 	& \textbf{0.08}	& \textbf{0.88} \\
  \bottomrule
  \end{tabular}
  }
  \vspace{-0.1cm}
  \caption{{\bf Quantitative comparison on Thuman2.0. }}
  \label{tb:quantity_comparison_thuman} 

\end{table}

\begin{table}[t]
  
      \renewcommand\arraystretch{0.8}

  \centering
  \resizebox{1.0\linewidth}{!}{
  \begin{tabular}{l| ccc c}
  \toprule

Method   &  PSNR $\uparrow$ & SSIM $\uparrow$	& LPIPS $\downarrow$ & CLIP-Similarity $\uparrow$  \\

  \midrule
    PIFu \cite{saito2019pifu}                 & 18.82	  & 0.91 & \textbf{0.08}	  & 0.82\\
    PAMIR \cite{zheng2021pamir}               & 18.06	  & 0.91 & \textbf{0.08}	  & 0.82\\
    SiTH \cite{ho2024sith}                 & 15.87	  & 0.89 & 0.10	  & 0.80\\
    SIFU \cite{zhang2024sifu}               & 15.90	  & 0.90 & 0.10	  & 0.83\\
    One2345 \cite{liu2024one1}                & 15.34	  & 0.84 & 0.16	  & 0.79\\
    SV3D \cite{voleti2024sv3d}                & 17.76	  & 0.87 & 0.13	  & 0.84\\

  \midrule
  \textbf{MVD-HuGaS (Ours)}   & \textbf{19.20}	& \textbf{0.92}  & \textbf{0.08}	& \textbf{0.90}\\
  \bottomrule
  \end{tabular}
  }
   \vspace{-0.1cm}
  \caption{{\bf Quantitative comparison on 2K2K. }}
  \label{tb:quantity_comparison_2k2k} 

  \vspace{-0.4cm}
\end{table}

\begin{figure}[t]
  \vspace{-0.4cm}
  \begin{center}
     \includegraphics[trim={0cm 0.3cm 0cm 0.1cm},clip,width=1.0 \linewidth]{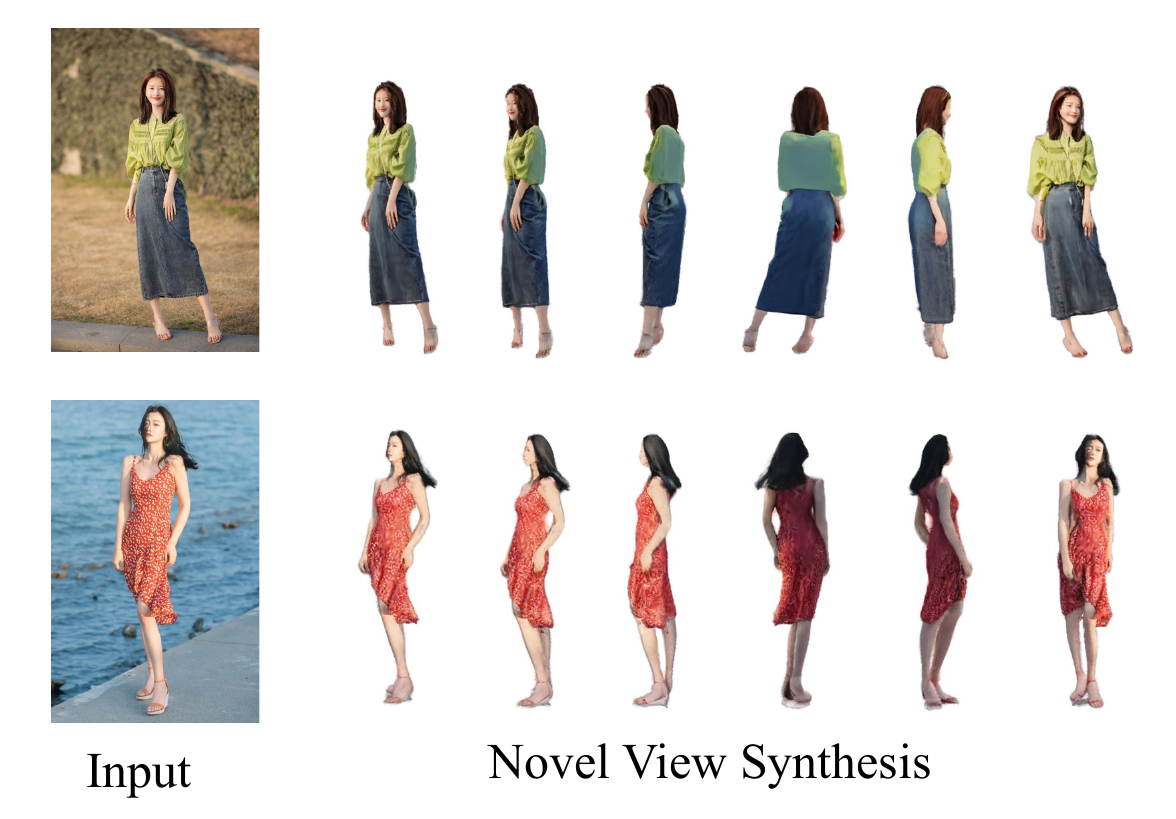}
  \end{center}
  \vspace{-0.4cm}
  \caption{\textbf{Qualitative results on wild images.}}
  \label{fig:wild}
  \vspace{-0.4cm}
\end{figure}

\begin{figure*}[h!]
  \vspace{-0.5cm}
  \begin{center}
     \includegraphics[trim={0cm 0cm 0cm 0.3cm},clip,width=0.85\linewidth]{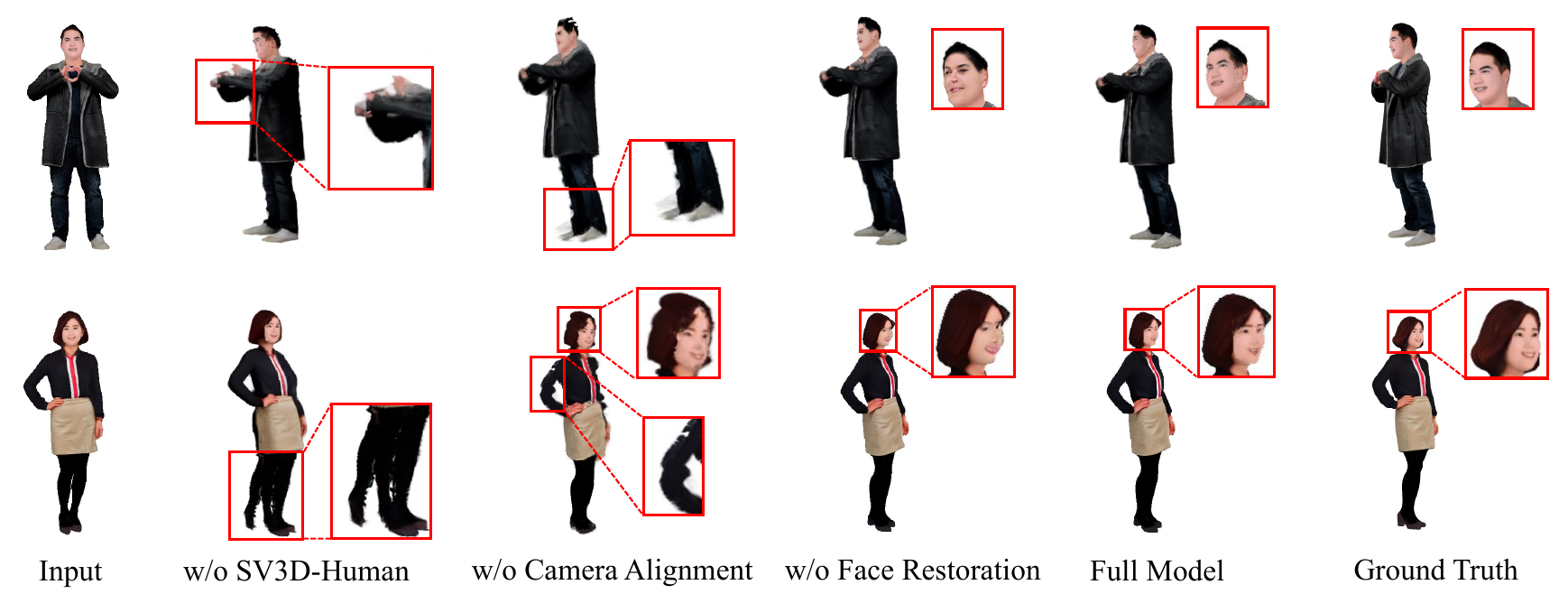}
  \end{center}
  \vspace{-0.5cm}
  \caption{\textbf{Ablation on each component in MVD-HuGaS.} Please zoom in to see the details.}
  \label{fig:ablation}
  \vspace{-0.4cm}
\end{figure*}

\subsection{Datasets}

\noindent \textbf{Thuman2.0 Dataset} captures 525 textured 3D scans from a DSLR rig. 
Being a standard for testing clothed human reconstruction, it provides 3D models and textures. 
To maintain fairness, we reserved it for testing, following~\cite{zhang2024sifu} by evaluating on 21 selected scans.

\noindent \textbf{2K2K Dataset } is a comprehensive 3D human dataset with 2050 high-quality scans, each annotated with a 3D model featuring colored vertices. 
The dataset covers a wide range of human identities, clothing styles, and hairstyles, offering excellent diversity for training diffusion models. 
We randomly segmented it into training (2,027 scans) and testing (23 scans) subsets, using the former for SV3D-Human model training and the latter for performance evaluation.

\subsection{Implementation Details}
\noindent  \textbf{Training}
We fine-tune SV3D-Human on the 2K2K training set with eight 40G A100 GPUs for 7 days, using DeepSpeed ZeRO Stage 2 to manage GPU memory demands.
The 3D mesh is rendered into $21\times 576\times 576$ RGB images around a central human, with azimuths evenly spaced and elevation angles randomly set between -30° and 30°. 
PyTorch3D facilitates this rendering. 
The frontal image is the input, with the other 20 images as labels for diffusion model training.

\noindent \textbf{Testing}
For models in the test set, we render 16 images at 0-degree elevation, evenly covering the 360-degree azimuthal range with PyTorch3D. 
These serve as the ground truth for novel view synthesis quality assessment. 
Mimicking real-world scenarios, reference images are randomly selected from -30 to 30 degrees in elevation. 
A single image is fed into MVD-HuGaS, and 16 images are rendered from the reconstructed 3D Gaussians to compare with the ground truth to measure performance.

\subsection{Metrics and Baseline Methods}
We quantitatively assess the rendering quality through PSNR, SSIM~\cite{wang2004image}, and LPIPS~\cite{zhang2018unreasonable}. 
For semantic similarity, CLIP-Similarity~\cite{radford2021learning} serves as the metric. 
The reconstructed meshes or Gaussians are rendered at 16 azimuths with 0-degree elevation and then compared to the ground truth.
For assessing single-view 3D clothed human reconstruction, the proposed MVD-HuGas is benchmarked against four SOTA single-view 3D human reconstruction methods: PIFu~\cite{saito2019pifu}, PaMIR~\cite{zheng2021pamir}, 
SiTH~\cite{ho2024sith}, and SIFU~\cite{zhang2024sifu} and two representative generic 3D generation methods: One2345 \cite{liu2024one} and SV3D \cite{voleti2024sv3d} .

\subsection{Evaluation}

We present qualitative results comparing our method with  SOTA techniques, as shown in \cref{fig:qualitative_comparison}. 
PIFu offers decent geometry but struggles with blurry textures, especially in occluded areas. 
PaMIR tends to generate artifacts outside the body and incomplete reconstructions, with particularly poor head reconstruction. 
SIFU and SiTH, incorporating diffusion model priors, improve texture detail.
However, they both assume orthogonal projections, limiting their robustness with non-zero elevation angles typical of real-world images. 
SIFU can introduce unrealistic textures on clothing, while SiTH produces a flat appearance on side views. 
In contrast, our approach delivers high-fidelity novel-view synthesis, capturing nuanced details such as clothing wrinkles, hair, and facial features across various perspectives.
Additionally, quantitative results in  \cref{tb:quantity_comparison_thuman}  and \cref{tb:quantity_comparison_2k2k} evaluate performance in reconstruction and generation quality, demonstrating our method's superiority over baselines.
Moreover, results on in-the-wild images demonstrate that our method effectively generalizes to real scenarios, as shown in \cref{fig:wild}. More comprehensive results are provided in the supplementary material.

\begin{table}
  
      \renewcommand\arraystretch{0.8}
  \centering

     \small

  \resizebox{1.\linewidth}{!}{
   \setlength{\tabcolsep}{1.1mm}{

  \begin{tabular}{l>{\small}cccc}

  \toprule
  Components & PSNR $\uparrow$ & SSIM $\uparrow$	& LPIPS $\downarrow$  & CLIP-Similarity $\uparrow$ \\
  \midrule
  w/o SV3D-Human            & 18.01      & 0.88	& 0.12 	& 0.84\\
  w/o Camera Alignment    & 18.42 	   & 0.89	& 0.10 	& 0.86\\
  w/o Face Distortion Mitigation      & 19.14 	   & 0.91	& \textbf{0.08} 	& 0.89\\

  \midrule
  \textbf{Full Model} & \textbf{19.20} 	& \textbf{0.92}	& \textbf{0.08} 	& \textbf{0.90}\\
  \bottomrule
\end{tabular}}}
  \vspace{-0.2cm}
  \caption{{\bf Ablation study results on 2K2K.}}
  \label{tb:ablation_study}
  \vspace{-0.3cm}
\end{table}

\subsection{Ablation Study}

We performed ablation studies to evaluate the impact of individual components in our framework. 
By sequentially excluding each key component, we systematically measured their influence on performance, with results detailed in \cref{tb:ablation_study}. 
Additionally, visual comparisons in \cref{fig:ablation} provide a qualitative assessment, highlighting the individual contributions of each component to the overall effectiveness.

\noindent \textbf{SV3D-Human}
In a comparative experiment, we replaced SV3D-Human with the original SV3D to highlight its importance in our framework. 
The vanilla SV3D often generates images with unrealistic human structures, such as extra limbs in \cref{fig:ablation}. In contrast, SV3D-Human produces more plausible multi-view images. 
Without SV3D-Human, there is a significant performance drop in the final rendering \cref{tb:ablation_study}, with PSNR decreasing from 19.20 to 18.01, confirming the effectiveness of our enhanced model.

\noindent \textbf{Camera Alignment Module}
We evaluated the effect of the camera alignment module in our framework by introducing errors in the camera poses, which are totally unknown in wild images. 
This led to noticeable blurring, loss of detail, and an increase in artifacts, as shown in \cref{fig:ablation}. 
Consequently, there was a measurable performance degradation in metrics \cref{tb:ablation_study}, with PSNR dropping from 19.20 to 18.42, underscoring the importance of accurate poses.

\noindent \textbf{Face Distortion Mitigation}
We tested the necessity of face distortion mitigation by disabling it in our pipeline for comparative experiments. 
While quantitative metrics showed minimal degradation, as detailed in \cref{tb:ablation_study}, this is attributed to the relatively small facial area's impact on overall calculations.
Qualitatively, however, the absence of this module often results in distorted facial renderings, as depicted in \cref{fig:ablation}. 
Such distortions can significantly affect the visual appeal and practical application of the synthesized views. 
In contrast, introducing face distortion mitigation allows our model to produce more realistic facial profiles, enhancing both the aesthetic and practical value of the images.

\section{Conclusion}
We presented MVD-HuGaS, an effective approach for single-view human reconstruction, leveraging multi-view diffusion priors to enhance the robustness and generalizability of wild input images. 
We first injected 3D human priors into the multi-view diffusion model to get SV3D-Human, which enables the generation of reasonable multi-view human images from a single reference image.
To solve the camera misalignment problem, a camera optimization module was proposed to jointly optimize the camera parameters and 3D Gaussians via image reconstruction loss.
Furthermore, we introduced a depth-based face restoration module to combine the human face geometry knowledge with the input texture information, inpainting the distorted generated faces.
Experimental results validated the effectiveness of the proposed MVD-HuGaS on the Thuman2.0 and 2K2K.

{\small

\input{Main.bbl}
}

\end{document}